\documentclass[10pt,twocolumn,letterpaper]{article}

\usepackage[pagenumbers]{cvpr}
\usepackage{graphicx}
\usepackage{amsmath,amssymb}
\usepackage{booktabs}
\usepackage{multirow}
\usepackage{xcolor}
\usepackage{tikz}
\usetikzlibrary{positioning, arrows.meta, shapes.geometric, fit, calc, backgrounds, decorations.pathreplacing, patterns}
\usepackage{pgfplots}
\pgfplotsset{compat=1.18}
\usepackage{subcaption}
\usepackage[hidelinks]{hyperref}
\usepackage{enumitem}
\usepackage{balance}
\usepackage{algorithm}
\usepackage{algorithmic}

\newcommand{\method}{DART\xspace}

\definecolor{trtblue}{HTML}{76B900}
\definecolor{backbonecolor}{HTML}{4A90D9}
\definecolor{encdeccolor}{HTML}{E8833A}
\definecolor{textcolor}{HTML}{7B68EE}
\definecolor{postcolor}{HTML}{50C878}
\definecolor{prepcolor}{HTML}{9B59B6}
\definecolor{basegray}{HTML}{8899AA}
\definecolor{redundantred}{HTML}{D94444}

\begin{document}

\title{Detect Anything in Real Time:\\From Single-Prompt Segmentation to Multi-Class Detection}

\author{
  Mehmet Kerem Turkcan \\
  \textit{Columbia University} \\
  {\tt\small mkt2126@columbia.edu}
}

\maketitle

\begin{abstract}
Recent advances in vision-language modeling have produced promptable detection and segmentation systems that accept arbitrary natural language queries at inference time.
Among these, SAM3~\cite{sam3} achieves state-of-the-art accuracy by combining a ViT-H/14 backbone with cross-modal transformer decoding and learned object queries.
However, SAM3 processes a single text prompt per forward pass. Detecting $N$ categories requires $N$ independent executions, each dominated by the 439M-parameter backbone.

We present Detect Anything in Real Time (\method), a training-free framework that converts SAM3 into a real-time multi-class detector by exploiting a structural invariant: the visual backbone is \emph{class-agnostic}, producing image features independent of the text prompt.
This allows the backbone computation to be shared between all classes, reducing its cost from $O(N)$ to $O(1)$.
Combined with batched multi-class decoding, detection-only inference, and TensorRT FP16 deployment, these optimizations yield \textbf{5.6$\times$} cumulative speedup at 3 classes, scaling to \textbf{25$\times$} at 80 classes, without modifying any model weight.
On COCO val2017 (5{,}000 images, 80 classes), \method achieves \textbf{55.8 AP} at \textbf{15.8\,FPS} (4 classes, 1008$\times$1008) on a single RTX~4080, surpassing purpose-built open-vocabulary detectors trained on millions of box annotations.
For extreme latency targets, adapter distillation with a frozen encoder-decoder achieves \textbf{38.7 AP} with a 13.9\,ms backbone.
Code and models are available at \url{https://github.com/mkturkcan/DART}.
\end{abstract}

\section{Introduction}
\label{sec:intro}

Recent vision-language models~\cite{kirillov2023sam,ravi2024sam2} can detect and segment objects from natural language prompts with impressive fidelity.
SAM3~\cite{sam3} represents the current state of the art: given a text prompt and an image, it localizes and segments all matching instances.
The architecture comprises a 439M-parameter ViT-H/14 backbone with 32 blocks of windowed and global attention, a contrastive text encoder that maps class names to 256-dimensional token sequences, a 6+6-layer transformer encoder-decoder with cross-modal attention, and 200 learned object queries with per-query box regression and presence prediction heads.

These same properties make SAM3 expensive to deploy as a multi-class detector.
The model processes one text prompt per forward pass.
Detecting $N$ categories requires $N$ full executions, each dominated by the backbone (78\% of compute, 87 of 112\,ms on an RTX~4080), prohibiting real-time inference beyond a few classes.

We target the training-free setting.
SAM3's training data and infrastructure are large in scale, making retraining infeasible without equivalent resources.
When the backbone must be replaced for extreme latency targets, adapter distillation (training only a small feature projection layer while keeping the encoder-decoder frozen) preserves 69\% of ViT-H quality (38.7 vs.\ 55.8 AP), whereas concurrent full-pipeline distillation retains only 10\% (5.5 AP; Section~\ref{sec:distill_comparison}).

The key observation is that \textit{the ViT-H backbone is class-agnostic}: it processes only the image, so its features are independent of the text prompt.
This invariant, together with cacheable text embeddings and the decoder's ability to accept batched prompts, renders nearly all of the per-class overhead redundant.

We exploit these invariants through a hierarchy of training-free optimizations:
backbone sharing, batched multi-class decoding, detection-only inference, TensorRT FP16 deployment with graph restructuring, and inter-frame pipelining.
Applied cumulatively, these yield 5.6$\times$ speedup at 3 classes, scaling to over 25$\times$ at 80 classes, all without retraining.

Our contributions are:
\begin{itemize}[nosep,leftmargin=*]
  \item A \textbf{training-free optimization hierarchy} that exploits the class-agnostic backbone invariant to convert SAM3 into a real-time multi-class detector, achieving 5.6$\times$ speedup at 3 classes with $O(1)$ backbone cost.
  \item \textbf{Adapter distillation analysis} showing that freezing the encoder-decoder and training only a lightweight FPN adapter (38.7 AP) substantially outperforms full-pipeline distillation~\cite{zeng2025efficientsam3} on the same backbone, suggesting that preserving the original detection pipeline is important when the training data is inaccessible.
  \item \textbf{Deployment analysis} covering split-engine TRT design for open vocabulary, attention graph restructuring for correct FP16 deployment, and sub-block pruning at ViT sub-block granularity.
\end{itemize}

\section{Related Work}
\label{sec:related}

\paragraph{Open-vocabulary detection.}
OWL-ViT~\cite{minderer2022owlvit} adapts a CLIP backbone for open-vocabulary detection via query-based classification.
Grounding DINO~\cite{liu2023grounding} fuses text and image features through cross-modal attention in a DETR-like architecture.
YOLO-World~\cite{cheng2024yoloworld} integrates CLIP embeddings into a single-stage detector for real-time inference.
Unlike these purpose-built detectors, we show that an existing promptable model can be structurally repurposed for the same task without architectural changes or retraining.

\paragraph{Efficient ViT deployment.}
TensorRT~\cite{tensorrt} accelerates neural networks through graph fusion and kernel autotuning.
PyTorch~2.0 compilation~\cite{ansel2024pytorch2} provides an alternative via cuBLAS with FP32 accumulation.
We show that for deep ViTs, the ONNX graph structure determines whether TRT dispatches attention to accumulation-safe fused kernels or to generic FP16 matrix multiplications, and provide a restructuring that enables correct FP16 deployment at lower latency than PyTorch compilation.

\paragraph{Efficient SAM3 variants.}
Concurrent with our work, EfficientSAM3~\cite{zeng2025efficientsam3} proposes Progressive Hierarchical Distillation (PHD) to transfer SAM3's capabilities to lightweight backbones (RepViT~\cite{wang2024repvit}, TinyViT~\cite{wu2022tinyvit}, EfficientViT~\cite{cai2023efficientvit}).
PHD replaces the ViT-H backbone \emph{and} fine-tunes the encoder-decoder end-to-end, following the prompt-in-the-loop paradigm of EdgeSAM~\cite{zhou2023edgesam}.
Their evaluation measures instance segmentation quality (mask mIoU given ground-truth boxes), not open-vocabulary detection.
We show in Section~\ref{sec:distill_comparison} that end-to-end distillation substantially degrades detection quality compared to preserving the original encoder-decoder.

\paragraph{Training-free pruning.}
BlockPruner~\cite{zhong2024blockpruner} removes transformer blocks based on reconstruction loss, originally targeting LLMs.
We adapt it to vision transformers with sub-block granularity, decomposing each ViT block into independently removable attention and MLP sub-blocks.

\section{Method}
\label{sec:method}

Given a pre-trained vision-language model $\mathcal{M}$ that processes one text prompt per forward pass and a set of $N$ class names, the goal is to detect all instances of all $N$ classes in an input image at real-time latency, without modifying any model weight.

\begin{figure*}[t]
\centering
\begin{tikzpicture}[
    arr/.style={
      -{Stealth[length=4pt, width=3pt]}, line width=0.5pt, color=#1!60!black,
    },
  ]

  \def\rowh{0.50}
  \def\rowgap{0.15}
  \def\gap{0.08}
  \def\rr{2pt}
  \def\lw{0.5pt}

  \def\bbw{4.2}
  \def\edw{1.1}
  \def\sgw{0.6}
  \pgfmathsetmacro{\atotalw}{\bbw+\gap+\edw+\gap+\sgw}

  \foreach \i/\clabel in {0/Class 1, 1/Class 2, 2/Class 3} {
    \pgfmathsetmacro{\ybot}{2.6 - \i*(\rowh+\rowgap)}

    \fill[basegray!12, rounded corners=\rr]
      (0, \ybot) rectangle (\bbw, \ybot+\rowh);
    \draw[basegray!50!black, line width=\lw, rounded corners=\rr]
      (0, \ybot) rectangle (\bbw, \ybot+\rowh);
    \node[font=\scriptsize\sffamily, color=basegray!70!black]
      at (0.5*\bbw, \ybot+0.5*\rowh) {ViT-H Backbone};

    \pgfmathsetmacro{\edx}{\bbw+\gap}
    \fill[basegray!8, rounded corners=\rr]
      (\edx, \ybot) rectangle (\edx+\edw, \ybot+\rowh);
    \draw[basegray!50!black, line width=\lw, rounded corners=\rr]
      (\edx, \ybot) rectangle (\edx+\edw, \ybot+\rowh);
    \node[font=\scriptsize\sffamily, color=basegray!70!black]
      at (\edx+0.5*\edw, \ybot+0.5*\rowh) {Enc-Dec};

    \pgfmathsetmacro{\sgx}{\bbw+\gap+\edw+\gap}
    \fill[basegray!5, rounded corners=\rr]
      (\sgx, \ybot) rectangle (\sgx+\sgw, \ybot+\rowh);
    \draw[basegray!50!black, line width=\lw, rounded corners=\rr]
      (\sgx, \ybot) rectangle (\sgx+\sgw, \ybot+\rowh);
    \node[font=\tiny\sffamily, color=basegray!60!black]
      at (\sgx+0.5*\sgw, \ybot+0.5*\rowh) {Seg};

    \node[font=\tiny\sffamily, color=basegray!60!black, anchor=east]
      at (-0.15, \ybot+0.5*\rowh) {\clabel};
  }

  \draw[redundantred!70, line width=0.8pt, densely dashed, rounded corners=3pt]
    (-0.06, 2.6+\rowh+0.06) rectangle (\bbw+0.06, 2.6-2*\rowh-2*\rowgap-0.06);
  \node[font=\scriptsize\sffamily\bfseries, color=redundantred!80!black, anchor=south]
    at (0.5*\bbw, 2.6+\rowh+0.10) {$\times N$ redundant};

  \pgfmathsetmacro{\abracy}{2.6-2*(\rowh+\rowgap)-0.18}
  \draw[decorate, decoration={brace, amplitude=4pt, mirror}, gray!50, line width=0.4pt]
    (0, \abracy) -- (\atotalw, \abracy);
  \node[font=\tiny\sffamily, color=gray!60!black, anchor=north]
    at (0.5*\atotalw, \abracy-0.10) {$N \times 112$\,ms per image};

  \pgfmathsetmacro{\apctbracy}{\abracy-0.32}
  \draw[decorate, decoration={brace, amplitude=3pt, mirror}, basegray!40, line width=0.3pt]
    (0, \apctbracy) -- (\bbw, \apctbracy);
  \node[font=\tiny\sffamily, color=basegray!50!black, anchor=north]
    at (0.5*\bbw, \apctbracy-0.08) {78\% of compute};

  \node[font=\scriptsize\sffamily\bfseries, anchor=north]
    at (0.5*\atotalw, \apctbracy-0.35) {(a) SAM3: per-class inference};

  \pgfmathsetmacro{\arrowx}{\atotalw + 0.35}
  \draw[-{Stealth[length=6pt, width=4pt]}, line width=1.2pt, color=backbonecolor!60!black]
    (\arrowx, 2.20) -- ({\arrowx+1.15}, 2.20);
  \node[font=\tiny\sffamily, color=backbonecolor!70!black, align=left, anchor=south west]
    at (\arrowx-0.05, 2.35) {%
      share backbone\\
      batch $N$ classes\\
      remove masks\\
      $\rightarrow$ TRT FP16};

  \pgfmathsetmacro{\bstart}{\arrowx + 1.3}

  \def\bkw{2.8}    %
  \def\enw{2.4}    %
  \def\nmw{0.5}    %
  \def\prow{1.95}  %

  \pgfmathsetmacro{\bkx}{\bstart}
  \fill[backbonecolor!12, rounded corners=\rr]
    (\bkx, \prow) rectangle ({\bkx+\bkw}, \prow+\rowh);
  \draw[backbonecolor!50!black, line width=\lw, rounded corners=\rr]
    (\bkx, \prow) rectangle ({\bkx+\bkw}, \prow+\rowh);
  \node[font=\scriptsize\sffamily, color=backbonecolor!70!black]
    at ({\bkx+0.5*\bkw}, \prow+0.5*\rowh) {ViT-H Backbone};

  \node[font=\scriptsize\sffamily\bfseries, color=backbonecolor!70!black, anchor=south east]
    at ({\bkx+\bkw-0.05}, \prow+\rowh+0.02) {$1\times$};

  \pgfmathsetmacro{\enx}{\bkx+\bkw+0.5}
  \fill[encdeccolor!12, rounded corners=\rr]
    (\enx, \prow) rectangle ({\enx+\enw}, \prow+\rowh);
  \draw[encdeccolor!50!black, line width=\lw, rounded corners=\rr]
    (\enx, \prow) rectangle ({\enx+\enw}, \prow+\rowh);
  \node[font=\scriptsize\sffamily, color=encdeccolor!70!black]
    at ({\enx+0.5*\enw}, \prow+0.5*\rowh) {Batched Enc-Dec};

  \pgfmathsetmacro{\nmx}{\enx+\enw+0.3}
  \fill[postcolor!12, rounded corners=\rr]
    (\nmx, \prow) rectangle ({\nmx+\nmw}, \prow+\rowh);
  \draw[postcolor!50!black, line width=\lw, rounded corners=\rr]
    (\nmx, \prow) rectangle ({\nmx+\nmw}, \prow+\rowh);
  \node[font=\tiny\sffamily, color=postcolor!70!black]
    at ({\nmx+0.5*\nmw}, \prow+0.5*\rowh) {NMS};

  \draw[arr=backbonecolor] ({\bkx+\bkw}, \prow+0.5*\rowh) --
    node[above, font=\tiny\sffamily, color=gray!55!black] {FPN}
    (\enx, \prow+0.5*\rowh);
  \draw[arr=encdeccolor] ({\enx+\enw}, \prow+0.5*\rowh) -- (\nmx, \prow+0.5*\rowh);

  \pgfmathsetmacro{\texty}{\prow+\rowh+0.55}
  \fill[textcolor!12, rounded corners=\rr]
    (\enx, \texty) rectangle ({\enx+\enw}, {\texty+0.40});
  \draw[textcolor!50!black, line width=\lw, rounded corners=\rr]
    (\enx, \texty) rectangle ({\enx+\enw}, {\texty+0.40});
  \node[font=\tiny\sffamily, color=textcolor!70!black]
    at ({\enx+0.5*\enw}, {\texty+0.20}) {Text Encoder (cached)};
  \draw[arr=textcolor] ({\enx+0.5*\enw}, \texty) --
    node[right, font=\tiny\sffamily, color=gray!55!black, pos=0.4] {$N$}
    ({\enx+0.5*\enw}, \prow+\rowh);

  \pgfmathsetmacro{\tmy}{\prow-0.08}
  \node[font=\tiny\sffamily\bfseries, color=backbonecolor!70!black, anchor=north]
    at ({\bkx+0.5*\bkw}, \tmy) {53\,ms};
  \node[font=\tiny\sffamily\bfseries, color=encdeccolor!70!black, anchor=north]
    at ({\enx+0.5*\enw}, \tmy) {7--41\,ms};
  \node[font=\tiny\sffamily\bfseries, color=postcolor!70!black, anchor=north]
    at ({\nmx+0.5*\nmw}, \tmy) {$<\!1$\,ms};

  \node[font=\tiny\sffamily\itshape, color=gray!55!black, anchor=north]
    at ({\bkx+0.5*\bkw}, {\tmy-0.22}) {explicit attention, real-valued RoPE};
  \node[font=\tiny\sffamily\itshape, color=gray!55!black, anchor=north]
    at ({\enx+0.5*\enw}, {\tmy-0.22}) {mask head removed};

  \pgfmathsetmacro{\brightx}{\nmx+\nmw}
  \pgfmathsetmacro{\bcenterx}{0.5*(\bstart+\brightx)}
  \node[font=\scriptsize\sffamily\bfseries, anchor=north]
    at (\bcenterx, \apctbracy-0.35) {(b) DART: shared backbone + batched decoding};

  \def\tscale{0.085}  %
  \def\yseq{-0.60}
  \def\ypipe{-1.45}
  \def\bblen{53}
  \def\edlen{17}

  \node[font=\scriptsize\sffamily, color=gray!60!black, anchor=east]
    at (-0.2, \yseq) {Sequential};

  \fill[backbonecolor!15, rounded corners=\rr]
    (0, \yseq-0.22) rectangle ({\bblen*\tscale}, \yseq+0.22);
  \draw[backbonecolor!45, line width=\lw, rounded corners=\rr]
    (0, \yseq-0.22) rectangle ({\bblen*\tscale}, \yseq+0.22);
  \node[font=\tiny\sffamily, color=backbonecolor!65!black]
    at ({0.5*\bblen*\tscale}, \yseq) {Backbone~53\,ms};

  \fill[encdeccolor!15, rounded corners=\rr]
    ({\bblen*\tscale}, \yseq-0.22) rectangle ({(\bblen+\edlen)*\tscale}, \yseq+0.22);
  \draw[encdeccolor!45, line width=\lw, rounded corners=\rr]
    ({\bblen*\tscale}, \yseq-0.22) rectangle ({(\bblen+\edlen)*\tscale}, \yseq+0.22);
  \node[font=\tiny\sffamily, color=encdeccolor!65!black]
    at ({(\bblen+0.5*\edlen)*\tscale}, \yseq) {E-D};

  \draw[gray!45, line width=0.4pt, {Stealth[length=2pt]}-{Stealth[length=2pt]}]
    (0, \yseq+0.34) -- ({(\bblen+\edlen)*\tscale}, \yseq+0.34)
    node[midway, above, font=\tiny\sffamily, gray!55!black] {70\,ms~=~14.3\,FPS};

  \node[font=\scriptsize\sffamily, color=gray!60!black, anchor=east]
    at (-0.2, \ypipe) {Pipelined};

  \fill[backbonecolor!15, rounded corners=\rr]
    (0, \ypipe-0.22) rectangle ({\bblen*\tscale}, \ypipe+0.22);
  \draw[backbonecolor!45, line width=\lw, rounded corners=\rr]
    (0, \ypipe-0.22) rectangle ({\bblen*\tscale}, \ypipe+0.22);
  \node[font=\tiny\sffamily, color=backbonecolor!65!black]
    at ({0.5*\bblen*\tscale}, \ypipe) {BB$(t)$~53\,ms};

  \fill[encdeccolor!15, rounded corners=\rr]
    ({\bblen*\tscale}, \ypipe-0.22) rectangle ({(\bblen+\edlen)*\tscale}, \ypipe+0.22);
  \draw[encdeccolor!45, line width=\lw, rounded corners=\rr]
    ({\bblen*\tscale}, \ypipe-0.22) rectangle ({(\bblen+\edlen)*\tscale}, \ypipe+0.22);
  \node[font=\tiny\sffamily, color=encdeccolor!65!black]
    at ({(\bblen+0.5*\edlen)*\tscale}, \ypipe) {E-D$(t)$};

  \fill[backbonecolor!15, rounded corners=\rr]
    ({\bblen*\tscale}, \ypipe-0.58) rectangle ({2*\bblen*\tscale}, \ypipe-0.98);
  \draw[backbonecolor!45, line width=\lw, rounded corners=\rr]
    ({\bblen*\tscale}, \ypipe-0.58) rectangle ({2*\bblen*\tscale}, \ypipe-0.98);
  \node[font=\tiny\sffamily, color=backbonecolor!45!black]
    at ({1.5*\bblen*\tscale}, \ypipe-0.78) {BB$(t{+}1)$};

  \fill[encdeccolor!15, rounded corners=\rr]
    ({2*\bblen*\tscale}, \ypipe-0.58) rectangle ({(2*\bblen+\edlen)*\tscale}, \ypipe-0.98);
  \draw[encdeccolor!45, line width=\lw, rounded corners=\rr]
    ({2*\bblen*\tscale}, \ypipe-0.58) rectangle ({(2*\bblen+\edlen)*\tscale}, \ypipe-0.98);
  \node[font=\tiny\sffamily, color=encdeccolor!45!black]
    at ({(2*\bblen+0.5*\edlen)*\tscale}, \ypipe-0.78) {E-D};

  \fill[red!14, rounded corners=1pt]
    ({\bblen*\tscale}, \ypipe-0.58) rectangle ({(\bblen+\edlen)*\tscale}, \ypipe+0.22);
  \draw[red!80, densely dashed, line width=0.3pt, rounded corners=1pt]
    ({\bblen*\tscale}, \ypipe-0.58) rectangle ({(\bblen+\edlen)*\tscale}, \ypipe+0.22);

  \draw[gray!45, line width=0.4pt, {Stealth[length=2pt]}-{Stealth[length=2pt]}]
    (0, \ypipe+0.34) -- ({\bblen*\tscale}, \ypipe+0.34)
    node[midway, above, font=\tiny\sffamily, gray!55!black] {53\,ms/frame~=~18.9\,FPS};

  \node[font=\tiny\sffamily, color=red!35!black, anchor=north]
    at ({(\bblen+0.5*\edlen)*\tscale}, \ypipe-1.01) {overlap};

  \node[font=\scriptsize\sffamily\bfseries, anchor=north]
    at ({0.5*(\bblen+\edlen)*\tscale}, \ypipe-1.22) {(c) Frame pipelining};

\end{tikzpicture}
\caption{%
  \textbf{From SAM3 to \method.}
  \textbf{(a)}~SAM3 runs the full pipeline once per class.
  The backbone accounts for 78\% of compute and is repeated $N$ times despite being class-independent.
  \textbf{(b)}~\method shares backbone features across all classes, batches the encoder-decoder, removes the mask head, and deploys both stages as TRT FP16 engines.
  The backbone uses restructured attention (explicit operations, real-valued RoPE) to enable correct FP16 export (\S\ref{sec:graph}).
  \textbf{(c)}~For video, backbone and encoder-decoder run on separate CUDA streams: the encoder-decoder for frame $t$ overlaps with the backbone for frame $t{+}1$, reducing per-frame latency from 70\,ms to 53\,ms (4 classes, 1008px).
}
\label{fig:pipeline}
\end{figure*}
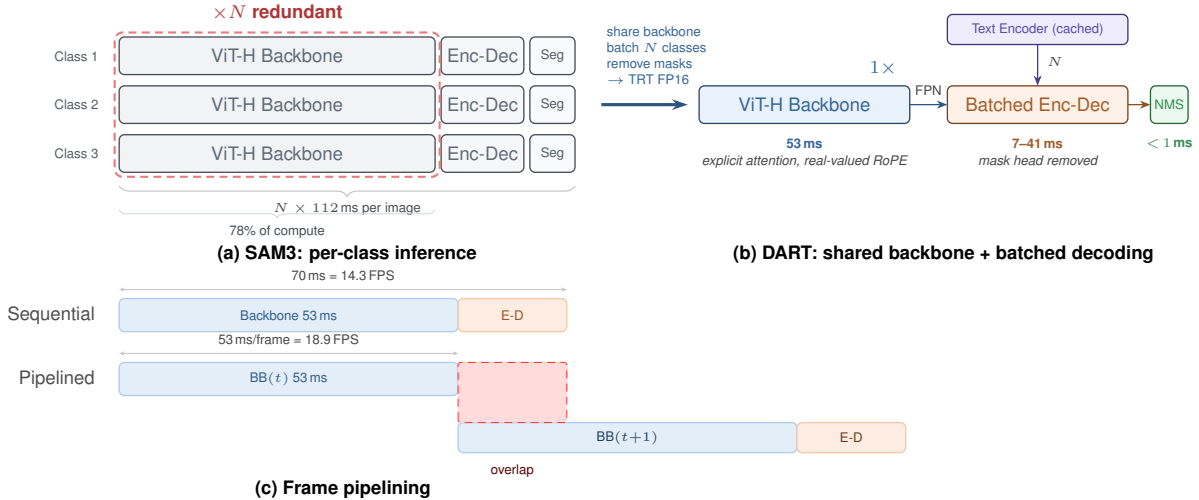

\subsection{SAM3 Cost Structure}
\label{sec:sam3_arch}

SAM3 processes an image-text pair through five stages:
(1)~a ViT-H backbone (439M parameters, 32 blocks with windowed/global attention and 2D RoPE) extracts three FPN feature levels,
(2)~a text encoder maps the class name to 32 tokens of dimension 256,
(3)~a 6-layer transformer encoder fuses image and text features via cross-attention over $S{=}5184$ spatial tokens,
(4)~a 6-layer decoder with 200 learned queries cross-attends to encoded features, and
(5)~dot-product scoring, a presence head, and per-query mask prediction produce final detections.

The backbone consumes 78\% of per-class latency (87 of 112\,ms on an RTX~4080), with the encoder-decoder, scoring heads, and mask prediction accounting for the remainder.

\subsection{Exploiting Structural Invariants}
\label{sec:multiclass}

Three structural properties of SAM3's architecture render the per-class overhead redundant, enabling its conversion from a per-class segmentation system into an efficient multi-class detector.

\paragraph{Backbone sharing.}
The backbone processes only pixel values.
We compute it once per image and reuse the FPN features for all $N$ classes, reducing backbone cost from $O(N)$ to $O(1)$.
For 3 classes, this alone provides a $2.1\times$ speedup (Table~\ref{tab:hierarchy}).

\paragraph{Batched multi-class decoding.}
A key structural property of the encoder-decoder is that text conditioning enters exclusively through cross-attention, which operates independently across the batch dimension.
All $N$ class prompts can therefore be stacked along the batch axis (text features $\mathbf{T} \in \mathbb{R}^{L_t \times N \times d}$ and image features $\mathbf{I} \in \mathbb{R}^{S \times N \times d}$) and processed in a single forward pass, converting $N$ sequential executions into one batched pass.

\paragraph{Detection-only inference.}
For applications requiring only bounding boxes, the mask prediction head can be removed entirely.
The decoder's learned presence token remains available for scoring class existence, and classes with low presence probability are filtered before NMS.

Together, these yield a $3.0\times$ speedup at 3 classes (Table~\ref{tab:hierarchy}).

\subsection{TensorRT FP16 Deployment}
\label{sec:graph}

The encoder-decoder exports to TensorRT FP16 without modification (25\,ms to 15\,ms for 3 classes).
The ViT-H backbone, however, requires attention graph restructuring.

Hardware inference compilers fuse multi-head attention (MHA) into kernels that internally accumulate in FP32, but rely on pattern matching to identify attention subgraphs.
Two common ViT constructs break this matching: complex-valued RoPE (no ONNX equivalent) and fused SDPA operators that decompose into non-canonical subgraphs.
When matching fails, the compiler falls back to generic FP16 matrix multiplications whose per-operation rounding errors (${\sim}10^{-4}$) compound through residual connections, producing catastrophically degraded features after 32 blocks (cosine similarity 0.058 vs.\ FP32; Table~\ref{tab:fp16}).
We restructure both operations at the PyTorch level: (1)~real-valued RoPE via precomputed $\cos\theta$/$\sin\theta$ buffers, and (2)~explicit $QK^\top\!\to\!$Scale$\to$Softmax$\to\!V$ attention.
These canonical forms restore pattern matching, yielding correct FP16 at 53\,ms (Table~\ref{tab:fp16}).

\subsection{Split-Engine Design for Open Vocabulary}
\label{sec:trt}

We decompose the pipeline into two independent TRT FP16 engines (backbone and encoder-decoder) that communicate through intermediate FPN feature tensors on GPU (Fig.~\ref{fig:pipeline}b).
The text encoder remains in PyTorch, computing embeddings once per class set and caching them as GPU tensors.
This split preserves the open-vocabulary property: changing detection categories requires only recomputing text embeddings (milliseconds), whereas a monolithic TRT export would require rebuilding the entire engine (minutes).
For large vocabularies (e.g., 80 COCO classes), the encoder-decoder processes up to $N_{\max}$ prompts per pass, requiring $\lceil N / N_{\max} \rceil$ passes when $N$ exceeds this limit.

\subsection{Inter-Frame Pipelining}
\label{sec:pipelining}

The split-engine design enables inter-frame pipelining for video. The backbone and encoder-decoder run on separate CUDA streams, so the backbone for frame $t{+}1$ overlaps with the encoder-decoder for frame $t$ (Fig.~\ref{fig:pipeline}c).
Per-frame latency is bounded by $\max(T_{\text{bb}}, T_{\text{ed}})$ instead of their sum:
\begin{equation}
\text{FPS}_{\text{pipe}} \leq \frac{1}{\max(T_{\text{bb}},\; T_{\text{ed}}(N))}
\label{eq:pipelining}
\end{equation}
In practice, scheduling overhead reduces this bound slightly (15.8 vs.\ 18.8\,FPS predicted at 4 classes, 1008px; Table~\ref{tab:class_scaling}).
The observed gains come from overlapping kernel-launch and data-movement phases rather than compute parallelism, as each engine already saturates the GPU's compute units.

\subsection{Training-Free Sub-Block Pruning}
\label{sec:pruning}

As an orthogonal strategy, we extend BlockPruner~\cite{zhong2024blockpruner} to ViT at \emph{sub-block granularity}: each of the 32 blocks is decomposed into independently removable attention and MLP sub-blocks, yielding 64 candidates.
A greedy search removes sub-blocks one at a time, selecting the one with minimal FPN feature reconstruction loss on calibration images.
The four global attention blocks (indices 7, 15, 23, 31) are protected as they provide the only cross-window information flow.
We denote a backbone with $K$ sub-blocks removed as \textbf{SBP-$K$}.
Algorithm~\ref{alg:pruning} details the procedure.

\begin{algorithm}[t]
\caption{Greedy sub-block pruning (SBP-$K$)}
\label{alg:pruning}
\small
\begin{algorithmic}[1]
\REQUIRE ViT backbone $\mathcal{B}$ with $L$ blocks, calibration images $\mathcal{X}$, budget $K$
\ENSURE Pruned backbone $\mathcal{B}'$
\STATE $\mathcal{C} \gets \{(\ell, \texttt{attn}), (\ell, \texttt{mlp}) \mid \ell = 1 \dots L\}$ \COMMENT{64 candidates}
\STATE Remove global-attention blocks from $\mathcal{C}$ \COMMENT{protect 7,15,23,31}
\STATE $\mathbf{F}^* \gets \text{FPN}(\mathcal{B}(\mathcal{X}))$ \COMMENT{reference features}
\FOR{$k = 1$ \TO $K$}
  \FOR{each candidate $c \in \mathcal{C}$}
    \STATE $\mathcal{B}_c \gets \mathcal{B}$ with sub-block $c$ removed
    \STATE $\delta_c \gets \|\text{FPN}(\mathcal{B}_c(\mathcal{X})) - \mathbf{F}^*\|_2$
  \ENDFOR
  \STATE $c^* \gets \arg\min_c \delta_c$
  \STATE Remove $c^*$ from $\mathcal{B}$ and $\mathcal{C}$
\ENDFOR
\RETURN $\mathcal{B}$
\end{algorithmic}
\end{algorithm}

\section{Experiments}
\label{sec:experiments}

\subsection{Setup}
\label{sec:setup}

We conduct all experiments on a single NVIDIA RTX~4080 (16\,GB VRAM) with PyTorch~2.7.0, TensorRT~10.9.0, and ONNX~1.20.1.
We use the SAM3 checkpoint with ViT-H/14 (439M backbone parameters).
We measure latency using traffic videos from the VisDrone-VID2019 dataset (100 frames after 10-frame warmup, averaged over 5 runs), and report detection quality as standard COCO box AP (averaged over IoU 0.50-0.95) on the full val2017 set (5{,}000~images, 80~classes) using the standard COCO evaluation protocol.

\subsection{Optimization Hierarchy}
\label{sec:ablation}

Table~\ref{tab:hierarchy} traces the cumulative speedup as each optimization level is applied.

\begin{table}[t]
\centering
\small
\caption{\textbf{Cumulative optimization hierarchy.} 1008$\times$1008, 3 classes, RTX~4080. Each row includes all optimizations from rows above.}
\label{tab:hierarchy}
\setlength{\tabcolsep}{3pt}
\begin{tabular}{@{}llcccc@{}}
\toprule
Level & Optimization & BB & E-D & ms & Speedup \\
\midrule
0 & Naive ($N$ full passes) & $N{\times}$ & $N{\times}$ & 336 & 1$\times$ \\
1 & Shared backbone & 1$\times$ & $N{\times}$ & 162 & 2.1$\times$ \\
2 & + Batched + det-only & 1$\times$ & 1$\times$ & 112 & 3.0$\times$ \\
3 & + Graph restructure + TRT BB & TRT & 1$\times$ & 78 & 4.3$\times$ \\
4 & + TRT enc-dec + pipeline & TRT & TRT & 60 & \textbf{5.6$\times$} \\
\bottomrule
\end{tabular}
\end{table}

Backbone sharing (Level~1) eliminates two of three backbone passes, cutting latency from 336\,ms to 162\,ms.
Batching all class prompts into a single encoder-decoder pass and removing the mask head (Level~2) reduces the remaining per-class overhead, reaching 112\,ms.
Graph restructuring enables correct TRT FP16 export of the backbone (Level~3, Section~\ref{sec:graph}), and deploying the encoder-decoder as a second TRT engine with inter-frame pipelining (Level~4) yields the final 60\,ms.
Each level builds on the previous; no single optimization alone achieves more than 2.1$\times$.
At 80 classes the cumulative speedup exceeds $25\times$, as backbone sharing amortizes 78\% of per-class cost across all categories.

\subsection{Scaling with Class Count}
\label{sec:class_scaling}

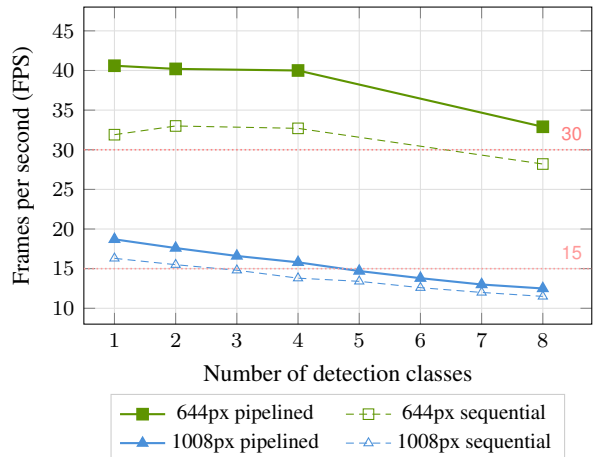
\begin{figure}[t]
\centering
\begin{tikzpicture}
\begin{axis}[
    width=\linewidth,
    height=5.8cm,
    xlabel={Number of detection classes},
    ylabel={Frames per second (FPS)},
    xmin=0.5, xmax=8.8,
    ymin=8, ymax=48,
    xtick={1,2,3,4,5,6,7,8},
    ytick={10,15,20,25,30,35,40,45},
    grid=both,
    grid style={gray!15, very thin},
    major grid style={gray!25, thin},
    legend style={
      at={(0.5,-0.22)}, anchor=north, legend columns=2,
      font=\footnotesize, draw=gray!40, fill=white,
      /tikz/every even column/.append style={column sep=8pt},
    },
    every axis plot/.append style={mark size=2pt},
    tick label style={font=\footnotesize},
    label style={font=\small},
  ]

  \addplot[color=trtblue!80!black, mark=square*, thick, solid]
    coordinates {(1,40.6) (2,40.2) (4,40.0) (8,32.9)};
  \addlegendentry{644px pipelined}

  \addplot[color=trtblue!80!black, mark=square, densely dashed, thin, mark options={solid}]
    coordinates {(1,31.9) (2,33.0) (4,32.7) (8,28.2)};
  \addlegendentry{644px sequential}

  \addplot[color=backbonecolor, mark=triangle*, thick, solid]
    coordinates {(1,18.7) (2,17.6) (3,16.6) (4,15.8) (5,14.7) (6,13.8) (7,13.0) (8,12.5)};
  \addlegendentry{1008px pipelined}

  \addplot[color=backbonecolor, mark=triangle, densely dashed, thin, mark options={solid}]
    coordinates {(1,16.3) (2,15.5) (3,14.8) (4,13.8) (5,13.4) (6,12.6) (7,12.0) (8,11.5)};
  \addlegendentry{1008px sequential}

  \draw[color=red!50, densely dotted, line width=0.6pt]
    (axis cs:0.5,30) -- (axis cs:8.8,30)
    node[pos=1, anchor=south east, font=\scriptsize\sffamily, color=red!50] {30};

  \draw[color=red!40, densely dotted, line width=0.6pt]
    (axis cs:0.5,15) -- (axis cs:8.8,15)
    node[pos=1, anchor=south east, font=\scriptsize\sffamily, color=red!40] {15};

\end{axis}
\end{tikzpicture}
\vspace{-2pt}
\caption{\textbf{FPS vs.\ class count.} At 644px, all tested class counts exceed 30\,FPS pipelined. At 1008px, up to 4 classes reach $\geq$15\,FPS. Pipelining provides 8-15\% improvement with diminishing returns as encoder-decoder cost approaches backbone cost.}
\label{fig:fps_vs_classes}
\end{figure}

Table~\ref{tab:class_scaling} and Figure~\ref{fig:fps_vs_classes} report throughput at both resolutions.
Real-time performance ($\geq$15\,FPS) is maintained for up to 4 classes at 1008px and all tested class counts at 644px.

\begin{table}[t]
\centering
\small
\caption{\textbf{Latency breakdown by class count} (1008$\times$1008, RTX~4080). Backbone latency is constant; encoder-decoder scales linearly. Pipelined FPS follows Eq.~\ref{eq:pipelining}.}
\label{tab:class_scaling}
\setlength{\tabcolsep}{4pt}
\begin{tabular}{@{}rcccccc@{}}
\toprule
$N$ & BB (ms) & E-D (ms) & Sum & Seq.\ FPS & Pipe.\ FPS & Gain \\
\midrule
1 & 53.2 & 7.9 & 61.1 & 16.3 & 18.7 & +15\% \\
2 & 53.2 & 11.4 & 64.6 & 15.5 & 17.6 & +14\% \\
4 & 53.2 & 19.2 & 72.4 & 13.8 & 15.8 & +14\% \\
8 & 53.2 & 34.7 & 87.9 & 11.5 & 12.5 & +9\% \\
\bottomrule
\end{tabular}
\end{table}

\subsection{Resolution and Quality Trade-offs}
\label{sec:resolution}

Table~\ref{tab:map} compares detection quality across resolutions and pruning configurations.

\begin{table*}[t]
\centering
\small
\caption{\textbf{COCO val2017 detection quality} (5{,}000 images, 80 classes). Resolution is the dominant quality lever, and small-object recall drives the gap.}
\label{tab:map}
\setlength{\tabcolsep}{6pt}
\begin{tabular}{@{}llccccccc@{}}
\toprule
Configuration & Res. & Removed & AP & AP$_{50}$ & AP$_{75}$ & AP$_S$ & AP$_M$ & AP$_L$ \\
\midrule
\textbf{Full TRT FP16} & \textbf{1008} & 0 & \textbf{55.8} & \textbf{73.4} & \textbf{61.5} & \textbf{40.3} & \textbf{59.8} & \textbf{70.7} \\
SBP-16 TRT FP16 & 1008 & 16 & 47.6 & 63.5 & 52.7 & 32.5 & 51.8 & 62.0 \\
\textbf{Full TRT FP16} & \textbf{644} & 0 & \textbf{39.1} & \textbf{63.9} & \textbf{39.9} & 12.4 & 44.4 & 65.4 \\
SBP-16 TRT FP16 & 644 & 16 & 32.8 & 54.5 & 33.4 & 9.9 & 37.9 & 56.7 \\
\bottomrule
\end{tabular}
\end{table*}

We find that reducing resolution from 1008 to 644 costs 30\% AP (55.8$\to$39.1), driven almost entirely by small-object recall (AP$_S$: 40.3$\to$12.4).
Large-object detection remains robust (AP$_L$: 70.7$\to$65.4).
SBP-16 pruning costs 15\% AP at 1008px with only 3\% latency reduction, because the protected global-attention blocks that dominate compute cannot be removed.
Resolution reduction is strictly more effective for speed: 644px at full depth (39.1~AP, 40~FPS) provides 2.7$\times$ the throughput of 1008px SBP-16 (47.6~AP, 15~FPS).

\begin{figure*}[t]
\centering
\begin{subfigure}[t]{0.325\textwidth}
  \centering
  \includegraphics[width=\linewidth]{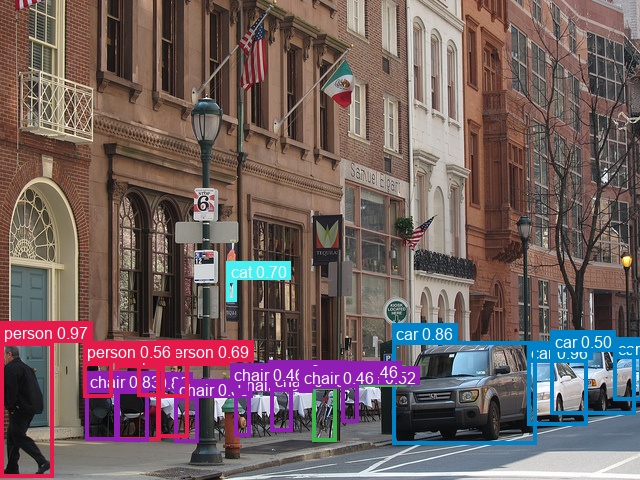}
\end{subfigure}\hfill
\begin{subfigure}[t]{0.325\textwidth}
  \centering
  \includegraphics[width=\linewidth]{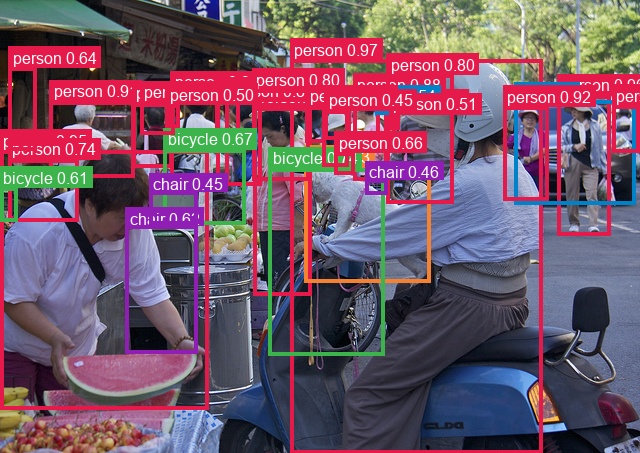}
\end{subfigure}\hfill
\begin{subfigure}[t]{0.325\textwidth}
  \centering
  \includegraphics[width=\linewidth]{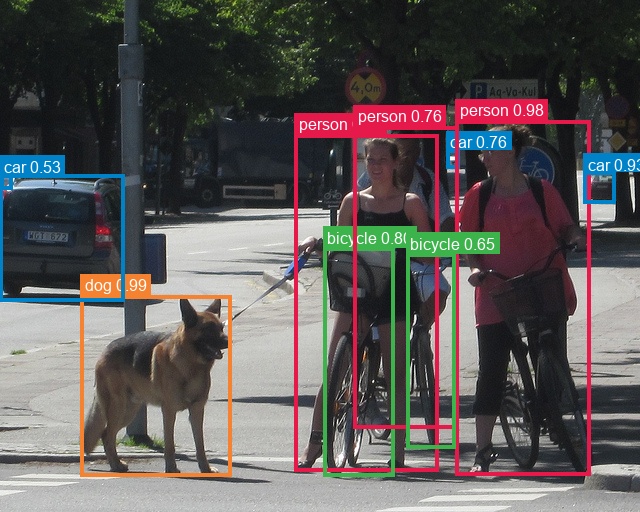}
\end{subfigure}
\caption{\textbf{Qualitative results.}
\method detections on COCO val2017 images using 6 open-vocabulary classes
(\textcolor{red}{person}, \textcolor{blue}{car}, \textcolor{orange}{dog},
\textcolor{green}{bicycle}, \textcolor{purple}{chair}, \textcolor{cyan}{cat})
at 1008px resolution with confidence threshold 0.45.
All optimizations are structural: \method produces identical outputs to the
per-class SAM3 baseline.}
\label{fig:qualitative}
\end{figure*}

\subsection{FP16 Precision Analysis}
\label{sec:fp16}

Deep residual vision transformers are vulnerable to FP16 accumulation error. Each generic FP16 matrix multiplication introduces a rounding error of ${\sim}10^{-4}$, and these errors compound through residual connections in all 32 blocks.
Because the error is distributed uniformly rather than localized to specific layers, selective per-block FP32 fallback is ineffective; only ensuring that \emph{every} attention layer dispatches to accumulation-safe fused kernels restores correctness.
Table~\ref{tab:fp16} validates this analysis: the fused-SDPA TRT path, which falls back to generic FP16 MatMul, produces catastrophically degraded features with a cosine similarity of 0.058.
Among the correct paths, explicit-attention graph restructuring (Section~\ref{sec:graph}) achieves the lowest latency.

\begin{table}[t]
\centering
\small
\caption{\textbf{Backbone deployment comparison.} 1008px, RTX~4080. Only the explicit-attention TRT path achieves both correctness and competitive latency.}
\label{tab:fp16}
\setlength{\tabcolsep}{4pt}
\begin{tabular}{@{}lccc@{}}
\toprule
Method & Latency & Cosine & Status \\
\midrule
\textbf{Explicit-attn TRT FP16} & \textbf{53\,ms} & \textbf{0.999} & \checkmark \\
Fused-SDPA TRT FP16 & 26\,ms & 0.058 & \texttimes \\
Fused-SDPA mixed (attn FP32) & 128\,ms & 0.999 & \checkmark \\
\texttt{torch.compile} FP16 & 75\,ms & 1.000 & \checkmark \\
PyTorch eager FP16 & 87\,ms & 1.000 & \checkmark \\
\bottomrule
\end{tabular}
\end{table}

\subsection{Backbone Distillation: Adapter vs.\ Full Pipeline}
\label{sec:distill_comparison}

When the backbone must be replaced for extreme latency targets, the feature distribution gap between student and ViT-H must be bridged.
We compare two strategies.

\smallskip\noindent\textbf{Adapter distillation (ours).}
A lightweight FPN adapter (${\sim}$5M parameters) projects student backbone features into the ViT-H feature space via multi-scale L2 matching.
The SAM3 encoder-decoder remains \emph{frozen} and only the adapter is trained.
We train on the COCO train2017 split (118K unlabeled images) for 5 epochs (${\sim}$2 GPU-hours on a single RTX~4080).

\smallskip\noindent\textbf{Full-pipeline distillation (EfficientSAM3-PHD~\cite{zeng2025efficientsam3}).}
This approach replaces the backbone and fine-tunes the encoder-decoder end-to-end via prompt-in-the-loop distillation on SA-1B.
The official EfficientSAM3 evaluation measures instance segmentation (mask mIoU given ground-truth boxes) rather than open-vocabulary detection; we re-evaluate their released checkpoints on the detection task for a direct comparison.
Crucially, the decoder weights are also modified.

\begin{table*}[t]
\centering
\small
\caption{\textbf{Backbone distillation comparison on COCO val2017} (80 classes, 1008$\times$1008, 5{,}000 images). Adapter distillation preserves the original SAM3 encoder-decoder, while full-pipeline distillation replaces it. EfficientSAM3 numbers are with presence-gating disabled. $^\dagger$TRT FP16 backbone + enc-dec ($N_{\max}{=}16$, 5 passes).}
\label{tab:distill}
\setlength{\tabcolsep}{5pt}
\begin{tabular}{@{}llccccccc@{}}
\toprule
Method & Backbone & Params & AP & AP$_{50}$ & AP$_{75}$ & AP$_S$ & AP$_L$ & BB (ms) \\
\midrule
\multicolumn{9}{@{}l}{\textit{Teacher (training-free, this work)}} \\
\method (ViT-H) & ViT-H/14 & 439M & \textbf{55.8} & \textbf{73.4} & \textbf{61.5} & \textbf{40.3} & \textbf{70.7} & 53 \\
\midrule
\multicolumn{9}{@{}l}{\textit{Adapter distillation (this work): frozen SAM3 encoder-decoder}} \\
\method (student)$^\ddagger$ & RepViT-M2.3 & 8.2M & \textbf{38.7} & \textbf{53.1} & \textbf{42.3} & \textbf{22.6} & \textbf{49.9} & 13.9 \\
\method (student)$^\ddagger$ & TinyViT-21M & 21M & 30.1 & 42.4 & 32.6 & 17.4 & 37.8 & 12.2 \\
\method (student)$^\ddagger$ & EfficientViT-L2 & 9.2M & 21.7 & 31.5 & 23.5 & 13.7 & 24.2 & 10.7 \\
\method (student)$^\ddagger$ & EfficientViT-L1 & 5.3M & 16.3 & 24.2 & 17.4 & 10.6 & 17.3 & \textbf{10.4} \\
\midrule
\multicolumn{9}{@{}l}{\textit{Full-pipeline distillation (EfficientSAM3-PHD~\cite{zeng2025efficientsam3})}} \\
ES-RV-L & RepViT-M2.3 & 8.2M & 5.5 & 7.9 & -- & 3.6 & 9.4 & -- \\
ES-TV-M & TinyViT-11M & 11M & 4.3 & 6.3 & -- & 4.5 & 7.4 & -- \\
ES-EV-M & EfficientViT-B1 & 4.8M & 4.3 & 6.4 & -- & 4.5 & 7.4 & -- \\
\bottomrule
\end{tabular}
\end{table*}

With the same RepViT-M2.3 backbone, adapter distillation achieves 38.7 AP compared to 5.5 AP for EfficientSAM3's full-pipeline distillation (Table~\ref{tab:distill}).
Note that EfficientSAM3's 5.5 AP is measured with the presence-gating mechanism disabled; with it enabled (the default SAM3 scoring pipeline), their models produce near-zero detections (0.1 AP), indicating that end-to-end distillation disrupts the presence token without recovering it.
The gap is consistent across all backbones and size categories.
The adapter approach preserves the original encoder-decoder weights; the student features need only approximate the encoder's expected input distribution.
By contrast, full-pipeline distillation modifies the decoder without access to the original training data, breaking learned scoring mechanisms.
These results suggest that, when the original training distribution is inaccessible, preserving the encoder-decoder and adapting only the feature interface is preferable to end-to-end distillation.

Student backbones reach 45--51\,FPS pipelined at 4 classes, converging to ${\sim}$14\,FPS beyond 8 classes as the encoder-decoder becomes the bottleneck (Fig.~\ref{fig:pareto}).

\paragraph{Training cost.}
All our adapter models train in under 2 hours on a single RTX~4080 using only COCO train2017 (118K unlabeled images, 5 epochs).
This represents three orders of magnitude less compute than EfficientSAM3's full-pipeline distillation on SA-1B, yet produces substantially better detection quality.
The low cost makes adapter distillation practical for rapid deployment to new hardware or backbone architectures.

\section{Discussion}
\label{sec:discussion}

\paragraph{Generalization beyond SAM3.}
The class-agnostic backbone property exploited here is not specific to SAM3.
Any vision-language architecture that processes images independently of text prompts exposes the same invariant, including SAM2~\cite{ravi2024sam2} and Grounding DINO~\cite{liu2023grounding}.
The optimization hierarchy (backbone sharing, batched decoding, split-engine deployment) applies directly to these architectures; only the graph restructuring details may differ depending on the backbone's attention implementation.

\paragraph{Context among open-vocabulary detectors.}
Table~\ref{tab:ov_comparison} situates \method among purpose-built open-vocabulary detectors.
Despite using no detection training data, \method achieves 55.8 AP, surpassing Grounding DINO-L (52.5 AP) and GLIP-L (49.8 AP), both trained on Objects365+GoldG whose categories overlap substantially with COCO.
This suggests that SAM3's segmentation pre-training yields detection capabilities that transfer directly without detection-specific fine-tuning.

\begin{table}[t]
\centering
\small
\caption{\textbf{Context among open-vocabulary detectors} (COCO val2017, 80 classes). \method uses frozen SAM3 weights with no detection training; all other methods train on large-scale detection data. $^\dagger$Zero-shot transfer (no COCO images in training). $^\ddagger$LVIS zero-shot; COCO zero-shot not reported.}
\label{tab:ov_comparison}
\setlength{\tabcolsep}{3pt}
\begin{tabular}{@{}llccc@{}}
\toprule
Method & Backbone & Det.\ Data & AP \\
\midrule
\multicolumn{4}{@{}l}{\textit{Trained on detection data (zero-shot on COCO)$^\dagger$}} \\
GLIP-L~\cite{li2022glip} & Swin-L & O365+GoldG & 49.8 \\
Grounding DINO-T~\cite{liu2023grounding} & Swin-T & O365+GoldG & 48.1 \\
Grounding DINO-L~\cite{liu2023grounding} & Swin-L & O365+GoldG+Cap4M & 52.5 \\
YOLO-World-L~\cite{cheng2024yoloworld} & YOLOv8-L & O365+GoldG & 44.9 \\
YOLO-World-X~\cite{cheng2024yoloworld} & YOLOv8-X & O365+GoldG & 46.7 \\
\midrule
\multicolumn{4}{@{}l}{\textit{No detection training (this work)}} \\
\method (1008px) & ViT-H/14 & None & \textbf{55.8} \\
\method (644px) & ViT-H/14 & None & 39.1 \\
\bottomrule
\end{tabular}
\end{table}

\begin{figure}[t]
\centering
\begin{tikzpicture}
\begin{axis}[
    width=\linewidth,
    height=5.5cm,
    xlabel={Pipelined FPS},
    ylabel={COCO AP},
    xmin=5, xmax=60,
    ymin=0.08, ymax=0.62,
    grid=both,
    grid style={gray!15, very thin},
    major grid style={gray!25, thin},
    every axis plot/.append style={thick},
    legend style={
      at={(0.5,-0.22)}, anchor=north, legend columns=2,
      font=\footnotesize, draw=gray!40, fill=white,
      /tikz/every even column/.append style={column sep=8pt},
    },
    tick label style={font=\footnotesize},
    label style={font=\small},
  ]

  \addplot[only marks, color=backbonecolor, mark=square*, mark size=2.5pt]
    coordinates {(18.7, 0.558) (15.8, 0.558) (12.5, 0.558)};
  \addlegendentry{ViT-H 1008px}

  \addplot[only marks, color=trtblue!80!black, mark=triangle*, mark size=2.5pt]
    coordinates {(40.6, 0.391) (40.0, 0.391) (32.9, 0.391)};
  \addlegendentry{ViT-H 644px}

  \addplot[only marks, color=redundantred, mark=*, mark size=3pt]
    coordinates {(50.6, 0.163)};
  \addlegendentry{EfficientViT-L1}

  \addplot[only marks, color=encdeccolor, mark=square*, mark size=3pt]
    coordinates {(49.1, 0.217)};
  \addlegendentry{EfficientViT-L2}

  \addplot[only marks, color=prepcolor, mark=diamond*, mark size=3pt]
    coordinates {(46.7, 0.301)};
  \addlegendentry{TinyViT-21M}

  \addplot[only marks, color=trtblue!50!black, mark=triangle*, mark size=3pt]
    coordinates {(45.3, 0.387)};
  \addlegendentry{RepViT-M2.3}

  \node[font=\tiny\sffamily, anchor=south, yshift=2pt, color=backbonecolor!80!black] at (axis cs:18.7,0.558) {1};
  \node[font=\tiny\sffamily, anchor=south, yshift=2pt, color=backbonecolor!80!black] at (axis cs:15.8,0.558) {4};
  \node[font=\tiny\sffamily, anchor=south, yshift=2pt, color=backbonecolor!80!black] at (axis cs:12.5,0.558) {8};

  \node[font=\tiny\sffamily, anchor=south, yshift=2pt, color=trtblue!60!black] at (axis cs:40.6,0.391) {1};
  \node[font=\tiny\sffamily, anchor=south west, yshift=2pt, color=trtblue!60!black] at (axis cs:40.0,0.391) {4};
  \node[font=\tiny\sffamily, anchor=south, yshift=2pt, color=trtblue!60!black] at (axis cs:32.9,0.391) {8};

  \node[font=\tiny\sffamily, anchor=west, xshift=3pt] at (axis cs:50.6,0.163) {4};
  \node[font=\tiny\sffamily, anchor=west, xshift=3pt] at (axis cs:49.1,0.217) {4};
  \node[font=\tiny\sffamily, anchor=west, xshift=3pt] at (axis cs:46.7,0.301) {4};
  \node[font=\tiny\sffamily, anchor=west, xshift=3pt] at (axis cs:45.3,0.387) {4};

  \draw[color=gray!40, densely dotted, line width=0.5pt]
    (axis cs:15,0.08) -- (axis cs:15,0.50);
  \node[font=\tiny\sffamily, color=gray!50, rotate=90, anchor=south] at (axis cs:15.5,0.18) {15\,FPS};

  \draw[color=gray!40, densely dotted, line width=0.5pt]
    (axis cs:30,0.08) -- (axis cs:30,0.50);
  \node[font=\tiny\sffamily, color=gray!50, rotate=90, anchor=south] at (axis cs:30.5,0.18) {30\,FPS};

\end{axis}
\end{tikzpicture}
\vspace{-2pt}
\caption{\textbf{Speed-quality Pareto front.} Numbers indicate class count. Teacher ViT-H dominates quality (55.8 AP) but is limited to $<$19\,FPS. Distilled students reach 45--51\,FPS at 4 classes, trading AP for 2.5--3$\times$ throughput. Among student backbones, RepViT-M2.3 achieves the highest AP (38.7) at 45\,FPS.}
\label{fig:pareto}
\end{figure}
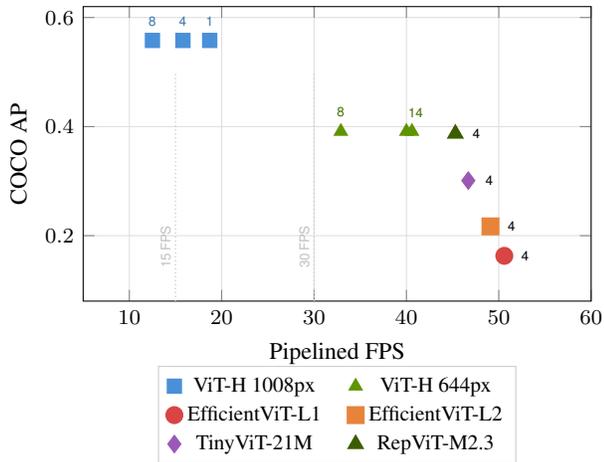

\paragraph{Encoder-decoder as the next bottleneck.}
For vocabularies beyond 16 classes, the encoder-decoder dominates total latency.
Encoder-decoder distillation, layer pruning, or prompt-batching strategies that reduce per-class decoding cost are natural next steps; each would extend the Pareto front in Figure~\ref{fig:pareto} toward higher class counts without sacrificing the frozen-pipeline advantage.

\paragraph{Limitations.}
Our approach inherits SAM3's single-scale FPN design, which limits small-object recall at lower resolutions (AP$_S$: 40.3$\to$12.4 from 1008 to 644\,px).
All benchmarks use a single RTX~4080; absolute latencies will differ on other hardware.

\section{Conclusion}
\label{sec:conclusion}

We have shown that SAM3 can be converted into a real-time open-vocabulary detector through training-free structural optimization, achieving 55.8 AP at 15.8\,FPS (4 classes) on COCO val2017, surpassing purpose-built detectors without any detection training.
Adapter distillation with a RepViT-M2.3 backbone provides a complementary path, reaching 38.7 AP at 45\,FPS.
Intermediate resolutions, encoder-decoder layer pruning, and adapter training on larger datasets remain promising directions for further improving the speed-quality trade-off.

\paragraph{Broader impact.}
By reducing the computational cost of open-vocabulary detection, \method lowers the barrier to deploying flexible object detection on commodity hardware.
This may benefit applications in assistive technology, autonomous navigation, and ecological monitoring, where real-time detection of user-specified categories is valuable.
However, the same capability also reduces the cost of deploying detection systems for surveillance.
Open-vocabulary detection is particularly sensitive in this regard, as it allows arbitrary target categories to be specified without retraining, potentially enabling misuse that would be more difficult with fixed-category detectors.
We encourage practitioners to consider privacy and consent frameworks before deployment in public spaces.

\section*{Acknowledgements}
This work was supported by the NSF Engineering Research Center for Smart Streetscapes under Award EEC-2133516.

{\small
\bibliographystyle{ieeenat_fullname}
\bibliography{refs}
}

\end{document}